\documentclass[
]{ceurart}

\usepackage{listings}
\pdfoutput=1

\begin{document}

\copyrightyear{2026}
\copyrightclause{Copyright for this paper by its authors.
  Use permitted under Creative Commons License Attribution 4.0
  International (CC BY 4.0).}

\conference{Joint Proceedings of REFSQ-2026 Workshops, Doctoral Symposium, Posters \& Tools Track, and Education and Training Track. Co-located with REFSQ 2026. Poznan, Poland, March 23-26, 2026}

\title{From Reviews to Requirements: Can LLMs Generate Human-Like User Stories?}

\author[1]{Shadman Sakib}[%
email=shadmansakib20@iut-dhaka.edu,
url=https://github.com/ShadmanSakib44/Review-to-Requirement/,
]

\fnmark[1]
\address[1]{Department of Computer Science and Engineering,
Islamic University of Technology (IUT), Gazipur, Bangladesh
}

\author[1]{Oishy Fatema Akhand}[%
email=oishyfatema@iut-dhaka.edu,
]
\fnmark[1]

\author[1]{Tasnia Tasneem}[%
email=tasniatasneem@iut-dhaka.edu,
]
\fnmark[1]

\author[1]{Shohel Ahmed}[%
orcid=0000-0001-8820-1672,
email=a.shohel@iut-dhaka.edu,
]
\cormark[1]

\cortext[1]{Corresponding author.}
\fntext[1]{These authors contributed equally.}

\begin{abstract}
  App store reviews provide a constant flow of real user feedback that can help improve software requirements. 
However, these reviews are often messy, informal, and difficult to analyze manually at scale. Although automated techniques exist, many do not perform well when replicated and often fail to produce clean, backlog-ready user stories for agile projects. In this study, we evaluate how well large language models (LLMs) such as GPT-3.5 Turbo, Gemini 2.0 Flash, and Mistral 7B Instruct 
can generate usable user stories directly from raw app reviews. Using the Mini-BAR dataset of 1,000+ health app reviews, we tested zero-shot, one-shot, and two-shot prompting methods. We evaluated the generated user stories using both human judgment (via the RUST framework) and a RoBERTa classifier fine-tuned on UStAI to assess their overall quality. Our results show that LLMs can match or even outperform humans in writing fluent, well-formatted user stories, especially when few-shot prompts are used. However, they still struggle to produce independent and unique user stories, which are essential for building a strong agile backlog. Overall, our findings show how LLMs can reliably turn unstructured app reviews into actionable software requirements, providing developers with clear guidance to turn user feedback into meaningful improvements.
\end{abstract}

\begin{keywords}
  User reviews\sep requirement engineering\sep large language models (LLMs)\sep user story 
\end{keywords}

\maketitle

\section{Introduction}

App stores like Google Play and the Apple App Store are a huge, ever-growing source of user feedback through millions of constantly evolving app reviews. These reviews reflect users’ experiences, expectations, and problems related to mobile applications and have been widely recognized as a valuable input for requirements engineering and software evolution activities ~\cite{1, 2}. In agile development contexts, such feedback can inform prioritization decisions and support the identification of new or evolving requirements. However, the scale and unstructured nature of app reviews, which are typically written in informal language and contain mixed concerns, make their systematic analysis challenging.

To address these challenges, prior research  ~\cite{3} has proposed a range of app review mining techniques, including sentiment analysis, topic modeling, and supervised classification for identifying feature requests, bug reports, and usability issues. While these approaches enable large-scale analysis of user feedback, they typically do not produce requirements artifacts that can be directly integrated into agile workflows. In practice, development teams rely on user stories as the primary representation of requirements in product backlogs. User stories are expected to follow a predefined structure and to be sufficiently clear, concise, and independent to support planning and implementation. Transforming raw app reviews into such backlog-ready user stories, therefore, is still a difficult and challenging job.

Recent advances in large language models (LLMs) have renewed interest in automating language-intensive requirements engineering tasks. LLMs have demonstrated strong capabilities in natural language understanding and generation and have recently been explored in requirements engineering related activities such as requirements refinement, analysis, and summarization ~\cite{6,7}. However, empirical evidence on their effectiveness in generating high-quality user stories directly from real-world, noisy app reviews is still limited. In particular, it is still unclear how user stories generated by large language models compare to those written by humans, how sensitive they are to different prompting strategies, or whether the model can even stick to a standard format like \textit{"As a… I want… so that…"} reliably every time.

This paper addresses these open questions through an empirical study on LLM-based user story generation from app reviews. Specifically, we investigate the following research questions:
\begin{itemize}
    \item \textbf{RQ1:} To what extent can LLMs preserve a predefined user story structure when transforming real-world app reviews into user stories?
    \item \textbf{RQ2:} How do different prompt strategies (zero-shot, one-shot, and two-shot) affect the quality of user stories generated by LLMs?
    \item \textbf{RQ3:} How does the quality of user stories generated by large language models compare with those authored by agile experts?
\end{itemize}

To answer these research questions, we empirically evaluate three large language models, GPT-3.5 Turbo, Gemini 2.0 Flash, and Mistral 7B Instruct, for generating user stories from app store reviews using zero-shot, one-shot, and two-shot prompting. User story quality is assessed through a multi-faceted evaluation combining human judgment using the RUST framework ~\cite{8}, automated classification with a RoBERTa ~\cite{25} model fine-tuned on UStAI Dataset ~\cite{9}. Our results show that LLMs can match, and in some cases outperform, agile experts in producing fluent and well-formatted user stories from app reviews, particularly when few-shot prompting is applied. Both human and automated evaluations indicate that LLM-generated user stories adhere closely to predefined templates and exhibit high readability. However, the models continue to struggle with generating independent and unique user stories, often producing semantically overlapping outputs from similar reviews. Since independence is essential for effective agile backlogs, these limitations highlight the need for additional backlog refinement support when using LLMs in requirements engineering.

The remainder of this paper is organized as follows: Section 2 reviews related work, and Section 3 details our methodology. Section 4 presents the results and discusses their practical impact. Finally, Section 5 outlines threats to validity, followed by conclusions in Section 6.

\section{Literature Review}
\label{sec:literature-review}

App store reviews are widely used in requirements engineering because they provide large-scale insights into bugs, feature requests, and user experience concerns reported by users ~\cite{10}.
Mapping and review studies show that topic modeling, sentiment analysis, and classifiers are commonly used to categorize app reviews into bugs, feature requests, and user feedback, but they do not produce backlog-ready items ~\cite{11,12}. Prior AI-based approaches mainly focus on evaluating or classifying existing user stories ~\cite{13,14} or generating them from structured inputs rather than noisy user feedback ~\cite{15,16}. Early studies with ChatGPT and other LLMs show that AI-generated user stories are readable and well structured, but often need human refinement, especially for scope and independence ~\cite{17,18}.

In requirements engineering more broadly, recent reviews highlight both the promise of LLMs for elicitation, refinement, and classification, and challenges around reliability, adherence to templates, and managing large, noisy corpora such as user reviews ~\cite{19,20}.
Large Language Models have fundamentally shifted the paradigm from extraction to generation. Wei et al.~\cite{21} demonstrated that LLMs can perform robust zero-shot classification and mining across English and Chinese reviews without task-specific training, addressing multilingual challenges in global app markets. However, their work remained focused on extraction rather than story generation. More recent systems have explored end-to-end generation: GeneUS~\cite{8} uses GPT-4 to produce user stories with test specifications from requirement documents, achieving strong readability but weaker specifiability in practitioner evaluations. UStAI~\cite{9} provides a benchmark of 3,000 LLM-generated stories from AI research abstracts, demonstrating that prompt design and post-generation repair significantly improve quality metrics. Santos et al.~\cite{22} compared ChatGPT with human-written outputs using INVEST criteria and found that LLMs perform well in fluency and readability but are weaker in completeness and actionable details. 



Quality evaluation frameworks have matured alongside generation methods. The Quality User Story (QUS) framework~\cite{23} establishes 13 criteria including atomicity, minimality, uniqueness, and independence as the de facto standard for assessing user story quality. The RUST framework~\cite{8} complements this with practitioner-focused dimensions: readability, understandability, specifiability, and technical adequacy. These frameworks reveal that LLM-generated stories often excel in surface-level quality but struggle with actionable precision required for implementation.

Although prior work has explored review mining and LLM-based requirement generation, directly converting raw, unstructured app reviews into high-quality user stories remains underexplored. Existing studies often rely on curated or structured inputs, avoiding the noise of real user feedback. Our work addresses this gap by using systematic prompting to generate backlog-ready user stories from raw app reviews, evaluated through both human judgment (RUST) and automated quality assessment based on QUS criteria.

\section{Methodology}
\label{sec:methodology}
\begin{figure}[ht]
  \centering
    \includegraphics[width=1.0\linewidth]{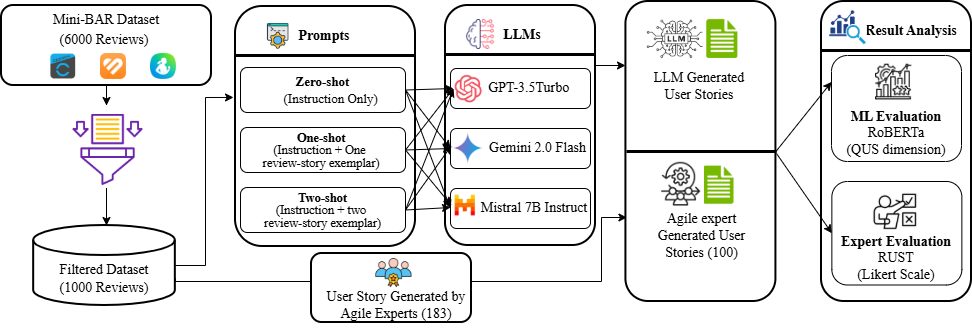}
    \caption{Overview of the proposed methodology.}
    \label{fig:overview}
\end{figure}

Figure~\ref{fig:overview} provides an overview of the proposed methodology for generating and evaluating user stories from mobile application reviews using large language models (LLMs). The approach consists of dataset preparation, prompt design, LLM configuration, and a multi-level evaluation setup that combines machine and human expert analysis.

We utilize the Mini-BAR corpus by Wei et al. ~\cite{21}, which comprises approximately 6,000 English-language reviews from the Garmin Connect, Huawei Health, and Samsung Health applications. Reviews are preprocessed at the sentence level by retaining emojis and symbols and discarding a review only if any sentence contains more emojis than tokens, in order to preserve expressive user feedback while removing noise. A length-based filter is subsequently applied to retain reviews ranging from 40 to 320 characters, ensuring sufficient context and readability. After preprocessing, a curated set of 1,000 reviews is selected for LLM-based user story generation. To establish a human baseline, two agile experts \footnote{\url{https://linkedin.com/in/nahidkamal-me/}} \footnote{\url{https://linkedin.com/in/jalalbd/}} were each given 100 app reviews and asked to write user stories based on these reviews. This process resulted in a total of 183 expert-written user stories. For the RUST-based human evaluation, 100 user stories are randomly selected from this set and used as the reference for quality assessment.






To investigate the effect of prompt design on user story generation, we employ three prompting strategies: zero-shot, one-shot, and two-shot prompting. In the zero-shot setting, LLMs are provided only with task instructions describing how to convert an app review into a user story. In the one-shot and two-shot settings, the instructions are augmented with one and two review–user story example pairs, respectively, to demonstrate the desired structure and level of abstraction. These prompting strategies are designed to address RQ2 by analyzing how increasing levels of contextual guidance influence the quality. We then evaluate three state-of-the-art LLMs: GPT-3.5 Turbo, Gemini 2.0 Flash, and Mistral 7B Instruct. Each model is applied to the filtered dataset under all three prompting strategies using default API parameters provided by each model. For each review, a single user story is generated without any task-specific hyperparameter tuning to ensure comparability across models and prompts. In total, this results in 9,000 LLM-generated user stories (1,000 reviews × 3 models × 3 prompting strategies).

We also employ a RoBERTa-based classifier to automatically assess user story quality. We use RoBERTa for automated evaluation because it is a discriminative model fine-tuned specifically for classification tasks, which makes it well-suited for evaluating multi-dimensional attributes of user stories. RoBERTa, an extension of BERT with dynamic masking and optimized training, provides strong natural language understanding capabilities and is well-suited for multi-label classification tasks. The model is fine-tuned on the UStAI dataset to predict QUS-aligned attributes, including atomicity, minimality, conceptual soundness, unambiguity, conflict-freeness, estimatability, independence, and uniqueness. The UStAI dataset contains approximately 3,000 user stories generated by three LLMs across 100 AI-system abstracts. Of these, 1,260 stories are manually annotated for QUS dimensions, non-functional requirements, and ethical considerations, providing a high-quality reference for training and evaluating the classifier.Model fine-tuning follows standard practice with binary cross-entropy loss, AdamW optimization (learning rate $2\times10^{-5}$), batch size 32, three epochs, linear learning-rate decay, and a maximum sequence length of 256. For each user story $s$, the model outputs per-criterion probabilities $P(y_i \mid s)$, which are aggregated into an overall quality score by averaging across all criteria. This aggregate score is used for comparative analysis across models and prompting strategies. Our automated evaluation enables a systematic comparison across models and prompting strategies and determines which LLMs perform best for high-quality user story generation. 

We also developed a web-based evaluation platform\footnote{\url{https://human-evaluation-website-t7sf.vercel.app}} to assess the quality of user stories and conducted an expert evaluation using the RUST  quality dimensions (Readability, Understandability, Specifiability, and Technical aspects) framework. We compared 100 agile expert-written user stories with user stories generated by the top three LLMs (identified through automated classification) on the same set of 100 reviews. The developed web interface presents one user review alongside four generated user stories from different sources, without disclosing their origins to the evaluators. For each story, evaluators rate all RUST dimensions using Likert-scale sliders and submit their assessment through a single submission button. Each submission is stored in the ratings table with a timestamp (\texttt{submitted\_at}), ensuring complete, traceable, and non-duplicated entries. The interface follows a minimalistic design with a single-task focus, automatic progression after submission, and persistent navigation, thereby reducing cognitive load and supporting consistent evaluation.

Finally, human evaluators were recruited through a professional freelancing platform\footnote{\url{https://www.upwork.com/jobs/~021967505800726342828}}. A total of 16 candidates responded to our call, from which three evaluators were selected through a structured screening process. The selection involved short interviews assessing familiarity with agile methodologies, experience with user stories, and professional background. As part of the screening, candidates were asked to convert sample user reviews into user stories and answer domain-specific questions. The final evaluators had 3–6 years of professional experience and represented diverse industry roles and all with prior agile experience. This process ensured that the human assessment was conducted by qualified practitioners with relevant domain knowledge. Table~\ref{tab:demo} summarizes the demographic and professional background of the human evaluators involved in the study.

\begin{table}[ht]
\scriptsize
\setlength{\tabcolsep}{3pt}
\caption{Demographic Information of Human Evaluators}
\label{tab:demo}
\begin{tabular}{llcl}
\toprule
\textbf{Evaluator} & \textbf{Role} & \textbf{Agile Experience} & \textbf{Region}  \\
\midrule
E1 & Senior Project Manager & 6+ & Europe \\
E2 & DevOps Engineer & 3+ & Asia \\
E3 & Systems Engineer & 3+ & Europe \\

\bottomrule
\end{tabular}
\end{table}

\section{Results and Discussion}
\label{ch:results_discussion}

Our analysis reveals the following results in response to the research questions.

\textbf{RQ1: To what extent can LLMs preserve a predefined user story structure when transforming real-world app reviews into user stories?}

\begin{figure}[ht]
\centering
\includegraphics[width=0.8\linewidth]{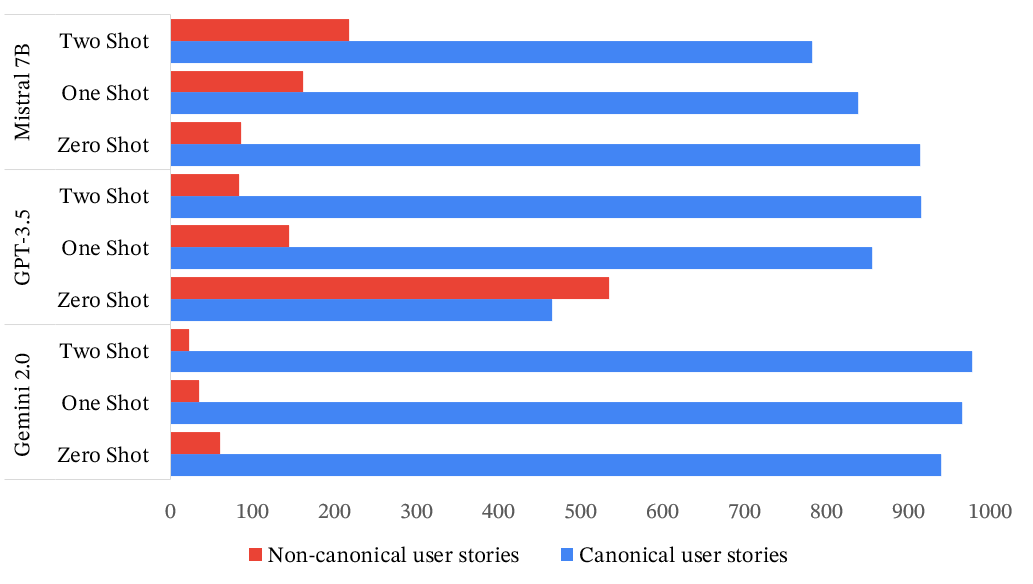}
\caption{Canonical and non-canonical user stories produced by different LLMs under zero-, one-, and two-shot prompting.}
\label{fig:RQ1}
\end{figure}

To investigate the extent to which LLMs preserve a predefined user story structure, one of the authors of this paper, who has over 10 years of industry experience and more than 6 years of Agile practice, manually reviewed all 9,000 LLM-generated user stories. Each story was evaluated against the canonical template ~\cite{24}: \texttt{As a [role], I want [feature] so that [benefit].} Figure~\ref{fig:RQ1} shows the distribution of user stories across different models and prompt strategies, distinguishing between stories that follow the canonical template and those that do not.

We found that, using a few examples (few-shot) generally helped models follow the correct structure better than giving no examples (zero-shot). Two-shot prompts slightly outperformed one-shot, suggesting that providing multiple examples gives models stronger guidance for following the canonical format. Gemini 2.0 showed exceptional structural consistency, producing very few non-canonical outputs even in zero-shot mode. Its adherence improved further with one- and two-shot prompts, showing it naturally aligns well with the target format and responds strongly to minimal guidance. GPT-3.5 Turbo also benefited significantly from few-shot prompting. While its zero-shot outputs often (about 50\%) deviated from the template, both one- and two-shot strategies reduced these deviations, with two-shot prompting achieving near-perfect adherence. This demonstrates GPT-3.5 Turbo’s ability to learn structural patterns effectively when provided with additional examples. However, Mistral 7B showed an opposite trend, performing best under zero-shot conditions, while adding one- or two-shot examples actually increased deviations from the canonical form. This suggests that for this smaller model, example-based prompts may introduce noise or cause overfitting to superficial features rather than capturing the underlying structure. This behavior contrasts with larger, instruction-tuned models and highlights the need for caution when applying standard few-shot strategies to smaller architectures.

\par\noindent\rule{\linewidth}{0.2pt}
{\textit{\textbf{Answer to the RQ1:} Using more examples (few-shot) usually helped models follow the canonical user story format better, but some models did well even without examples (zero-shot), showing that sensitivity to prompts varies across models. }}
\par\noindent\rule{\linewidth}{0.2pt}

\textbf{RQ2: How do different prompt strategies (zero-shot, one-shot, and two-shot) affect the quality of
user stories generated by LLMs?}

To answer RQ2, we used a fine-tuned RoBERTa-base model trained on UStAI to automatically evaluate nine QUS-based quality dimensions in user stories written by LLMs and agile experts. For each user story, an overall quality score was computed as the average probability across these criteria. Although the number of human-written stories (200) is smaller than the LLM-generated stories (1000), all reported values represent average criterion-level probabilities, which normalizes for sample size differences. Therefore, the comparison focuses on relative performance trends rather than absolute counts.

Table~\ref{tab:qus_quality} reports the average quality scores for each criterion across human-written stories and LLM-generated stories using zero-shot, one-shot, and two-shot prompting. 
\begin{table*}[htbp]
\centering
\scriptsize
\caption{Automated Quality Evaluation Scores Across Models and Prompt Strategies}
\label{tab:qus_quality}
\setlength{\tabcolsep}{2pt}
\renewcommand{\arraystretch}{1.2}
\begin{tabular}{lcccccccccc}
\toprule
\multirow{2}{*}{\textbf{Quality Criterion}} 
& \multirow{2}{*}{\textbf{Human}} 
& \multicolumn{3}{c}{\textbf{Gemini 2.0 Flash}} 
& \multicolumn{3}{c}{\textbf{GPT-3.5 Turbo}} 
& \multicolumn{3}{c}{\textbf{Mistral 7B Instruct}} \\
\cmidrule(lr){3-5} \cmidrule(lr){6-8} \cmidrule(lr){9-11}
&  & \textbf{Two-Shot} & \textbf{One-Shot} & \textbf{Zero-Shot}
& \textbf{Two-Shot} & \textbf{One-Shot} & \textbf{Zero-Shot}
& \textbf{Two-Shot} & \textbf{One-Shot} & \textbf{Zero-Shot} \\
\midrule
Atomic             & 0.76 & \textbf{0.83} & 0.82 & 0.82 & 0.82 & 0.82 & 0.82 & 0.82 & 0.82 & 0.82 \\
Minimal            & 0.83 & \textbf{0.94} & 0.94 & 0.94 & 0.94 & 0.94 & 0.94 & 0.94 & 0.94 & 0.94 \\
Conceptually Sound & 0.72 & 0.81 & 0.81 & 0.80 & \textbf{0.82} & 0.82 & 0.81 & 0.81 & 0.81 & 0.82 \\
Problem-Oriented   & 0.69 & 0.78 & 0.78 & 0.77 & \textbf{0.79} & 0.78 & 0.78 & 0.77 & 0.77 & 0.78 \\
Unambiguous        & 0.56 & 0.60 & 0.60 & 0.58 & 0.59 & 0.60 & 0.59 & 0.60 & \textbf{0.60} & 0.57 \\
Conflict-Free      & 0.76 & \textbf{0.82} & 0.82 & 0.81 & 0.81 & 0.82 & 0.81 & 0.82 & 0.82 & 0.81 \\
Estimatable        & 0.57 & 0.62 & 0.61 & 0.60 & 0.61 & 0.61 & 0.60 & 0.62 & \textbf{0.62} & 0.59 \\
Independent        & \textbf{0.32} & 0.28 & 0.27 & 0.27 & 0.27 & 0.27 & 0.26 & 0.28 & 0.29 & 0.26 \\
Unique             & \textbf{0.33} & 0.29 & 0.30 & 0.30 & 0.29 & 0.29 & 0.29 & 0.29 & 0.29 & 0.31 \\
\bottomrule

\end{tabular}
\vspace{2pt}
\parbox{\textwidth}{\footnotesize
\textit{Note:} Each value represents the average model output probability $P(y_i \mid s)$ for a given quality criterion, computed across all evaluated user stories. Values are rounded to two decimal places for readability. Bold values indicate the highest score within each row based on the original (unrounded) values.
}
\end{table*}

The results show that LLMs are highly competitive with human authors on several criteria related to clarity, structure, and linguistic adequacy. For example, for atomicity, LLM-generated stories achieve scores of approximately 0.82–0.83 under one-shot and two-shot prompting, compared to 0.76 for human-written stories. Similarly, for minimality, LLMs substantially outperform humans, with scores of approximately 0.93–0.94 versus 0.83 for human-authored stories. Comparable trends are observed for conceptual soundness, problem-orientation, unambiguity, and conflict-freeness, particularly when few-shot prompting is applied. However, a consistent performance gap is observed for the independence and uniqueness criteria. Human-written stories score notably higher on both dimensions, with average scores of 0.32 for independence and 0.33 for uniqueness, whereas LLM-generated stories remain lower, typically ranging between 0.26–0.28 for independence and 0.29–0.31 for uniqueness, regardless of the model or prompt strategy. This suggests that while LLMs are effective at producing fluent, well-structured, and contextually appropriate user stories, they tend to generate content that is more repetitive and less independent, which can negatively affect backlog diversity and modularity.

\begin{figure}[ht]
\centering
\includegraphics[width=0.6\linewidth]{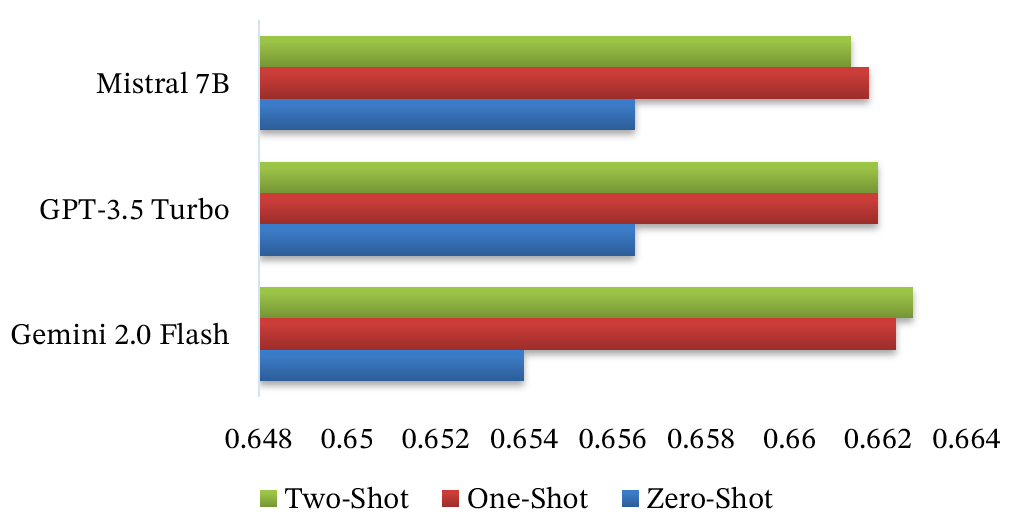}
\caption{Overall user-story quality scores for different LLMs under varying prompt strategies.}
\label{fig:RQ2}
\end{figure}

To provide a holistic view, we next compared the overall average quality scores across models and prompt strategies ans shown in Figure~\ref{fig:RQ2}. 

Across all three LLMs, few-shot prompting consistently outperforms zero-shot prompting. Zero-shot configurations achieve average quality scores around 0.654–0.657, while one-shot prompting raises performance to about 0.662, and two-shot prompting produces the highest scores, reaching $\approx0.663$. Differences between models are small (generally below 0.003), suggesting that prompt strategy has a stronger impact on user story quality than model choice. Overall, Gemini 2.0 Flash with two-shot prompting achieves the highest score (0.6628), with GPT-3.5 Turbo and Mistral 7B showing very similar performance under few-shot settings.

To statistically validate the observed differences, we further conducted a Kruskal-Wallis H test comparing the overall quality scores across the three prompt strategies. The test revealed a statistically significant effect of prompt strategy on user story quality ($\chi^2(2) = 1201.40, p < 0.0001$), indicating that at least one strategy differs significantly from the others. To examine the impact of prompt strategy on user story quality, we also conducted pairwise comparisons between Zero-Shot, One-Shot, and Two-Shot prompting strategies. As shown in Table ~\ref{tab:pairwise_comparison}. 
\begin{table}[htbp]
\centering
\caption{Pairwise Comparison of Prompting Strategies}
\label{tab:pairwise_comparison}
\begin{tabular}{lccccc}
\hline
\textbf{Sample1 - Sample2} & \textbf{Test Statistic} & \textbf{Std. Error} & \textbf{Std. Test Statistic ($z$)} & \textbf{Sig.} & \textbf{Adj. Sig.} \\
\hline
Zero-Shot vs One-Shot & -1,860.06 & 67.086 & -27.727 & $<$0.0001 & $<$0.0001 \\
Zero-Shot vs Two-Shot & -2,138.46 & 67.086 & -31.877 & $<$0.0001 & $<$0.0001 \\
One-Shot vs Two-Shot  & -278.396  & 67.086 & -4.15   & $<$0.0001 & $<$0.0001 \\
\hline
\end{tabular}
\end{table}
All pairwise differences were statistically significant (adjusted $p < 0.001$). Specifically, both One-Shot (Test Statistic = -1,860.06, $z = -27.727$) and Two-Shot (Test Statistic = -2,138.46, $z = -31.877$) strategies produced significantly higher quality scores than Zero-Shot prompting. Furthermore, Two-Shot prompting slightly outperformed One-Shot (Test Statistic = -278.396, $z = -4.15$), indicating that increasing the number of examples in the prompt enhances the overall quality of the generated user stories.

\par\noindent\rule{\linewidth}{0.2pt}
{\textit{\textbf{Answer to the RQ2:} Using few-shot prompts consistently improved user story quality, especially with two-shot prompting, though all models struggled to generate fully independent and unique stories. }}
\par\noindent\rule{\linewidth}{0.2pt}

\textbf{RQ3: How does the quality of user stories generated by large language models
compare with those authored by agile experts?}

To compare LLM-generated and expert-written user stories, we randomly selected 100 expert-written stories and generated the same stories using three LLM setups: Gemini 2.0 Two-Shot (highest quality in RQ2 and strong adherence to the canonical format from RQ1) and GPT-3.5 Turbo One- and Two-Shot (widely used and also following the canonical format). This resulted in 400 stories, which were evaluated by three human evaluators as described in Section 3. Table X shows the average RUST scores for each source. 
Gemini 2.0 Two-Shot achieved the highest overall score (Mean = 4.53, SD = 0.64), followed by GPT-3.5 Turbo Two-Shot (Mean = 4.49, SD = 0.59) and GPT-3.5 Turbo One-Shot (Mean = 4.45, SD = 0.64). Agile expert-written stories scored slightly lower (Mean = 4.39, SD = 0.70). These results indicate that top-performing LLMs can generate user stories of comparable or slightly higher quality than experienced human practitioners.

\begin{table}[htbp]
\centering
\caption{Overall RUST Evaluation Scores by Source}
\label{tab:overall_rust_scores}
\begin{tabular}{lccc}
\hline
\textbf{Source} & \textbf{Mean Score} & \textbf{Std. Dev.} & \textbf{N Ratings} \\
\hline
Gemini 2.0 Flash (Two-Shot) & 4.5265 & 0.6359 & 400 \\
GPT-3.5 Turbo (Two-Shot) & 4.4903 & 0.5949 & 400 \\
GPT-3.5 Turbo (One-Shot) & 4.4492 & 0.6384 & 400 \\
Agile Expert & 4.3879 & 0.6963 & 400 \\
\hline
\end{tabular}
\end{table}

\begin{figure}[ht]
\centering
\includegraphics[width=0.9\linewidth]{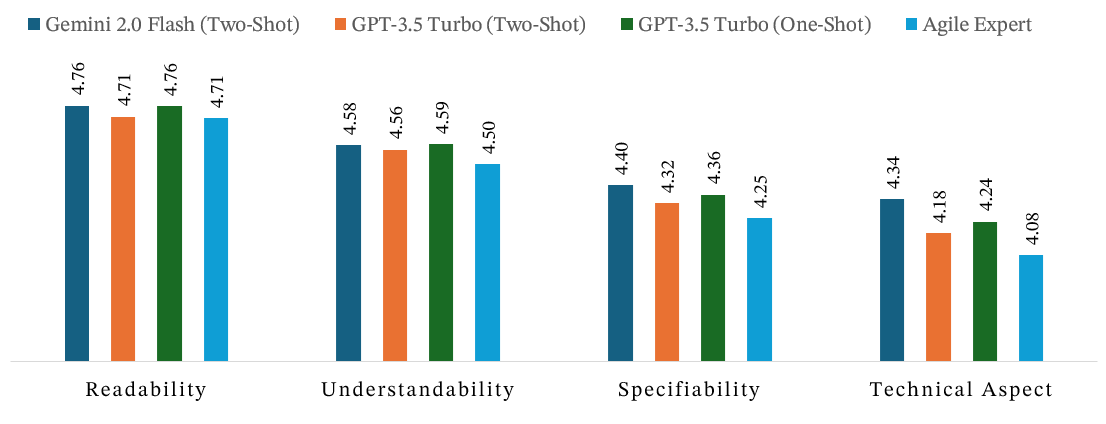}
\caption{Dimension-wise RUST evaluation scores by source.}
\label{fig:RQ3}
\end{figure}

We further analyzed the dimension-wise RUST evaluation scores (Figure~\ref{fig:RQ3}) understand performance across specific quality aspects. 

Across all dimensions, Gemini 2.0 Two-Shot consistently achieved the highest scores, while GPT-3.5 Turbo One- and Two-Shot performed slightly lower but generally exceeded expert-written stories. In readability, all LLMs scored above 4.71, with Gemini 2.0 Two-Shot reaching 4.76 versus 4.71 for experts, indicating fluent and easy-to-read stories. For understandability, LLMs outperformed experts (Gemini 4.58, GPT-3.5 Two-Shot 4.56, Expert 4.50), suggesting that the intended functionality is clearly communicated. Regarding specifiability, Gemini 2.0 Two-Shot scored 4.40, with GPT-3.5 One- and Two-Shot scoring 4.36 and 4.32, compared to 4.25 for experts, showing slightly more precise and actionable stories. Finally, for technical aspects, differences were smaller but present, with Gemini 2.0 Two-Shot at 4.34 versus 4.08 for experts, indicating technically coherent stories. All our results suggest that state-of-the-art LLMs not only match but can slightly exceed human performance in multiple quality dimensions. These findings highlight the potential of LLMs to support agile practitioners in generating high-quality user stories.

\par\noindent\rule{\linewidth}{0.2pt}
{\textit{\textbf{Answer to the RQ3:} State-of-the-art LLMs, particularly Gemini 2.0 Two-Shot, generate user stories of comparable or slightly higher quality than those written by agile experts across all RUST dimensions. }}
\par\noindent\rule{\linewidth}{0.2pt}

All our results show LLMs can save a lot of time in Agile teams. Developers and product owners can quickly turn hundreds or thousands of app reviews into good user stories using few-shot prompting with strong models like Gemini 2.0. This speeds up sprint planning and helps teams stay focused on real user needs.
However, to mitigate risks of repetitive or interdependent stories, practitioners should incorporate post-generation review mechanisms or hybrid approaches, such as LLM-assisted drafting followed by human refinement. Our work also links NLP and software engineering, adding to the growing field of AI in requirements engineering. It improves on past user story quality studies by mixing automatic scoring with human evaluation, and it clearly shows how much the choice of prompting style matters, especially depending on the model.

\section{Threats to Validity}
Following the framework proposed by Wohlin et al. \cite{26}, we organize the threats to validity in our study into four categories: internal validity, external validity, construct validity, and conclusion validity.
\subsection{Internal Validity}
In our study, human evaluation using the RUST framework was subjective. We tried to limit bias by assigning experienced agile practitioners and keeping the story sources anonymous, but personal judgment and past experience could still influence the results. Moreover, variations in prompt design and default LLM settings may have affected the generated stories that may impact the measured performance.

\subsection{External Validity}
Our used dataset was limited to 1,000 english reviews from three health-related applications, which may restrict generalizability to other languages, domains or LLM architectures. Since LLMs are trained on vast amounts of text that could include parts of our dataset, they might recognize some content. This means the results may not fully apply to new or unseen data.

\subsection{Construct Validity}
The automated evaluation utilizing the RoBERTa classifier and QUS-based criteria may not capture real-world aspects of a user story, like how easy it is to maintain, how well it fits into live backlogs, or how it affects team collaboration. Because of this, our metrics might not fully reflect the actual quality of the user stories.

\subsection{Conclusion Validity}
Differences in human judgments, LLM outputs, and evaluation methods could affect the results, that might make our conclusions less certain. While our study shows that LLMs can create high-quality user stories, human review is still important to make sure they work well in real agile development.

\section{Conclusion}

We investigated the ability of large language models to generate high-quality user stories directly from real-world app reviews. Three state-of-the-art LLMs (GPT-3.5 Turbo, Gemini 2.0 Flash, and Mistral 7B) were evaluated using zero-shot, one-shot, and two-shot prompting, with story quality assessed through automated metrics and expert human evaluation. Results show that LLMs, particularly under few-shot prompting, can produce user stories that are fluent, well-structured, and closely follow canonical templates. In many dimensions, including readability, understandability, and specifiability, LLM-generated stories match or slightly exceed those written by agile experts.

At the same time, challenges remain in independence and uniqueness, as LLMs often generate repetitive or overlapping stories. This indicates that human oversight is still important to ensure diverse, independent, and actionable requirements. Overall, our findings demonstrate that LLMs can complement human expertise in agile requirements engineering, providing scalable support for converting noisy app reviews into high-quality, actionable user stories. However, our evaluation was limited to three human raters and a single domain (health and fitness apps), future work should expand human assessment, explore cross-domain and cross-lingual generation, integrate LLMs with agile tools, and investigate advanced prompting strategies to handle more complex requirements.

\section*{Acknowledgements}
The authors would like to thank Md Shah Jalal, Senior Scrum Master at Cantaloupe Inc., and Nahid Kamal, Senior Technical Project Manager at BRAC IT Services Ltd., for their valuable contribution in writing 183 user stories from app reviews, which served as a human baseline in this study.

\section*{Declaration on Generative AI}
No AI tools were used to generate research results, perform analyses, or draw scientific conclusions. GPT-3.5 and Grammarly were used solely for language refinement, including paraphrasing, grammar and spelling correction, and style improvement. Turnitin was used to check textual similarity. After using these tools, the authors carefully reviewed and edited the content as necessary and take full responsibility for the publication’s content. 

\bibliography{sample-ceur}

@String{Computer = "{IEEE} Computer" }

@String{Springer = "Springer-Verlag" }

@article{1,
  title={Analysing app reviews for software engineering: a systematic literature review},
  author={D{\k{a}}browski, Jacek and Letier, Emmanuel and Perini, Anna and Susi, Angelo},
  journal={Empirical Software Engineering},
  volume={27},
  number={2},
  pages={43},
  year={2022},
  publisher={Springer}
}

@inproceedings{2,
  title={User feedback in the appstore: An empirical study},
  author={Pagano, Dennis and Maalej, Walid},
  booktitle={2013 21st IEEE international requirements engineering conference (RE)},
  pages={125--134},
  year={2013},
  organization={IEEE}
}

@article{3,
  title={From conventional methods to large language models: a systematic review of techniques in mobile app review analysis.},
  author={Arambepola, Nimasha and Munasinghe, Lankeshwara and Wimalasena, Waruni},
  journal={Interdisciplinary Journal of Information, Knowledge \& Management},
  volume={20},
  year={2025}
}

@article{6,
  title={Agile methodology for the standardization of engineering requirements using large language models},
  author={Tikayat Ray, Archana and Cole, Bjorn F and Pinon Fischer, Olivia J and Bhat, Anirudh Prabhakara and White, Ryan T and Mavris, Dimitri N},
  journal={Systems},
  volume={11},
  number={7},
  pages={352},
  year={2023},
  publisher={Multidisciplinary Digital Publishing Institute}
}

@incollection{7,
  title={Advancing requirements engineering through generative ai: Assessing the role of \uppercase{LLM}s},
  author={Arora, Chetan and Grundy, John and Abdelrazek, Mohamed},
  booktitle={Generative AI for Effective Software Development},
  pages={129--148},
  year={2024},
  publisher={Springer}
}

@article{8,
  title={Automated user story generation with test case specification using large language model},
  author={Rahman, Tajmilur and Zhu, Yuecai},
  journal={arXiv preprint arXiv:2404.01558},
  year={2024}
}

@inproceedings{9,
  title={Leveraging \uppercase{LLM}s for User Stories in AI Systems: UStAI Dataset},
  author={Yamani, Asma and Baslyman, Malak and Ahmed, Moataz},
  booktitle={Proceedings of the 21st International Conference on Predictive Models and Data Analytics in Software Engineering},
  pages={21--30},
  year={2025}
}

@inproceedings{10,
  title={Bug report, feature request, or simply praise? on automatically classifying app reviews},
  author={Maalej, Walid and Nabil, Hadeer},
  booktitle={2015 IEEE 23rd international requirements engineering conference (RE)},
  pages={116--125},
  year={2015},
  organization={IEEE}
}

@article{11,
  title={Mobile app review analysis for crowdsourcing of software requirements: a mapping study of automated and semi-automated tools},
  author={Massenon, Rhodes and Gambo, Ishaya and Ogundokun, Roseline Oluwaseun and Ogundepo, Ezekiel Adebayo and Srivastava, Sweta and Agarwal, Saurabh and Pak, Wooguil},
  journal={PeerJ Computer Science},
  volume={10},
  pages={e2401},
  year={2024},
  publisher={PeerJ Inc.}
}

@article{12,
  title={Evaluating pre-trained models for user feedback analysis in software engineering: A study on classification of app-reviews},
  author={Hadi, Mohammad A and Fard, Fatemeh H},
  journal={Empirical Software Engineering},
  volume={28},
  number={4},
  pages={88},
  year={2023},
  publisher={Springer}
}

@article{13,
  title={A requirement quality assessment method based on user stories},
  author={Xu, Xiangqian and Dou, Yajie and Qian, Liwei and Zhang, Zhiwei and Ma, Yufeng and Tan, Yuejin},
  journal={Electronics},
  volume={12},
  number={10},
  pages={2155},
  year={2023},
  publisher={MDPI}
}

@inproceedings{14,
  title={Classification of testable and valuable user stories by using supervised machine learning classifiers},
  author={Subedi, Ishan Mani and Singh, Maninder and Ramasamy, Vijayalakshmi and Walia, Gursimran Singh},
  booktitle={2021 IEEE International Symposium on Software Reliability Engineering Workshops (ISSREW)},
  pages={409--414},
  year={2021},
  organization={IEEE}
}

@inproceedings{15,
  title={Ai-generated user stories supporting human-centred development: an investigation on quality},
  author={Abed, Omed and Nebe, Karsten and Abdellatif, Ahmed Belal},
  booktitle={International Conference on Human-Computer Interaction},
  pages={3--13},
  year={2024},
  organization={Springer}
}

@inproceedings{16,
  title={A Two-Phase NLP-Driven Deep Learning Framework for Automated Software Requirements Classification and User Story Generation in Agile Development},
  author={Hany, Khaled M and Salama, Gouda I and Salem, Ahmed},
  booktitle={2025 International Conference on Machine Intelligence and Smart Innovation (ICMISI)},
  pages={177--182},
  year={2025},
  organization={IEEE}
}

@misc{17,
  title={Exploring \uppercase{LLM}s Impact on Student-Created User Stories and Acceptance Testing in Software Development. arxiv. org, 1401--1402},
  author={Brockenbrough, A and Feild, H and Salinas, D},
  year={2025}
}

@inproceedings{18,
  title={Using generative ai to create user stories in the software engineering classroom},
  author={Brockenbrough, Allan and Salinas, Dominic},
  booktitle={2024 36th International Conference on Software Engineering Education and Training (CSEE\&T)},
  pages={1--5},
  year={2024},
  organization={IEEE}
}

@article{19,
  title={Challenges in applying large language models to requirements engineering tasks},
  author={Norheim, Johannes J and Rebentisch, Eric and Xiao, Dekai and Draeger, Lorenz and Kerbrat, Alain and de Weck, Olivier L},
  journal={Design Science},
  volume={10},
  pages={e16},
  year={2024},
  publisher={Cambridge University Press}
}

@inproceedings{20,
  title={Leveraging Large Language Models for Requirements Generation: An Evaluation Through Systems Engineering Guidelines},
  author={Stein, Joel and Esho, Tomi and Gadewadikar, Jyotirmay},
  booktitle={2025 IEEE International Conference on AI and Data Analytics (ICAD)},
  pages={1--8},
  year={2025},
  organization={IEEE}
}

@inproceedings{21,
  title={Zero-shot bilingual app reviews mining with large language models},
  author={Wei, Jialiang and Courbis, Anne-Lise and Lambolais, Thomas and Xu, Binbin and Bernard, Pierre Louis and Dray, G{\'e}rard},
  booktitle={2023 IEEE 35th International Conference on Tools with Artificial Intelligence (ICTAI)},
  pages={898--904},
  year={2023},
  organization={IEEE}
}

@ARTICLE{22,
  title   = {User Stories: Does ChatGPT Do It Better?},
  year    = {2025},
  author  = {Santos, Reine and Freitas, Gabriel and Steinmacher, Igor and Conte, T. and Oran, Ana and Gadelha, Bruno},
  doi     = {10.5220/0013365500003929},
  journal = {International Conference on Enterprise Information Systems}
}

@article{23,
  title={Improving agile requirements: the quality user story framework and tool},
  author={Lucassen, Garm and Dalpiaz, Fabiano and van der Werf, Jan Martijn EM and Brinkkemper, Sjaak},
  journal={Requirements engineering},
  volume={21},
  number={3},
  pages={383--403},
  year={2016},
  publisher={Springer}
}

@article{24,
  title={Guide to the Software Engineering Body of Knowledge},
  author={Washizaki, Hironori},
  journal={IEEE Computer Society},
  year={2024}
}

@article{25,
  title={Roberta: A robustly optimized bert pretraining approach},
  author={Liu, Yinhan and Ott, Myle and Goyal, Naman and Du, Jingfei and Joshi, Mandar and Chen, Danqi and Levy, Omer and Lewis, Mike and Zettlemoyer, Luke and Stoyanov, Veselin},
  journal={arXiv preprint arXiv:1907.11692},
  year={2019}
}

@book{26,
  title={Experimentation in software engineering},
  author={Wohlin, Claes and Runeson, Per and H{\"o}st, Martin and Ohlsson, Magnus C and Regnell, Bj{\"o}rn and Wessl{\'e}n, Anders and others},
  volume={236},
  year={2012},
  publisher={Springer}
}

\end{document}